\documentclass[11pt]{article}
\usepackage{coling2020}
\usepackage{times}
\usepackage{url}
\usepackage{latexsym}
\usepackage{natbib}

\usepackage{times}
\usepackage{latexsym}
\usepackage{bm}
\usepackage{url}
\usepackage{amsmath}
\usepackage{amssymb}
\usepackage{pbox}
\usepackage{graphicx}
\usepackage{caption}
\usepackage{xspace}
\usepackage{multirow}

\graphicspath{ {./} }

\usepackage{color}
\usepackage{listings}
\usepackage{xcolor}
\usepackage{xspace}

\colingfinalcopy 

\title{Delexicalized Paraphrase Generation}

\author{Boya Yu \\
  Amazon Alexa AI \\
  {\tt boyayu@amazon.com} \\\And
  Konstantine Arkoudas \\
  Amazon Alexa AI \\
  {\tt arkoudk@amazon.com} \\\And
  Wael Hamza \\
  Amazon Alexa AI \\
  {\tt waelhamz@amazon.com}\\
}
\date{}

\begin{document}
\maketitle
\begin{abstract}
  We present a neural model for paraphrasing and train it to generate delexicalized sentences. We achieve this by creating training data in which each input is paired with a number of reference paraphrases. These sets of reference paraphrases represent a weak type of semantic equivalence based on annotated slots and intents. To understand semantics from different types of slots, other than anonymizing slots, we apply convolutional neural networks (CNN) prior to pooling on slot values and use pointers to locate slots in the output. We show empirically that the generated paraphrases  are of high quality, leading to an additional 1.29\% exact match on live utterances. We also show that natural language understanding (NLU) tasks, such as intent classification and named entity recognition, can benefit from data augmentation using automatically generated paraphrases.
\end{abstract}

\section{Introduction} \label{sec:intro}

\blfootnote{ 
 
 
 \hspace{-0.65cm} 
 This work is licensed under a Creative Commons 
 Attribution 4.0 International License. 
 License details: 
 \url{http://creativecommons.org/licenses/by/4.0/}. 
}

Paraphrases provide additional ways in which the same semantic meaning can be communicated through text or voice. Automatic paraphrase generation can benefit various
applications, including question answering \citep{fader2013paraphrase},
summarization \citep{barzilay2005sentence} and machine translation
\citep{callison2006improved,marton2009improved}.
Recently, neural paraphrasing methods have been proposed that utilize
sequence-to-sequence models \citep{prakash2016neural} or generative
models \citep{bowman2015generating,gupta2018deep}. Similar to other work \citep{sutskever2014sequence,mallinson2017paraphrasing}, we apply
an encoder-decoder model for paraphrasing, inspired by neural machine translation (NMT).

\textbf{Delexicalization} Unlike general paraphrases, which are typically reformulations of utterances, we paraphrase {\em delexicalized\/} sentences, in which named entities are replaced with generalized slot names. For example, ``I want to listen to \textit{Taylor Swift} 's \textit{Shake It Off}'' will be transformed into ``I want to listen to \{\textit{Artist}\}'s \{\textit{Music}\}.'' As a result, it is expected that the paraphrasing model will learn more about syntactic variations rather than semantic similarities among words.

An example application of our  paraphrasing model is third-party skill systems in digital voice assistants such as Amazon's Alexa. Users can extend Alexa's capabilities by ``skills.'' These skills are built by third-party developers, using the Alexa Skills Kit (ASK), and may cover any specific domain---Starbucks orders, Uber reservations, Jeopardy quizzes, and so on. Developers can build skills on the Alexa Developer Console, and start by defining an interactive model including an intent schema, slot types, sample utterances, and an invocation phrase \citep{kumar2017just}. The sample utterances can be delexicalized, and include general slots that can be filled by provided slot values. Sample JSON for a developer-defined skill can be found below. Our paraphrasing model generates delexicalized utterances that help developers create sample utterances for Alexa Skills, augmenting the training data of NLU (Natural Language Understanding) models and improving the performance of such models. 

\vspace{1.2cm}

  Sample JSON of ``play music'' skill:

\colorlet{punct}{red!60!black}
\definecolor{background}{HTML}{EEEEEE}
\definecolor{delim}{RGB}{20,105,176}
\colorlet{numb}{magenta!60!black}

\lstdefinelanguage{json}{
    basicstyle=\scriptsize\ttfamily,
    numbersep=8pt,
    showstringspaces=false,
    breaklines=true,
    frame=lines,
    backgroundcolor=\color{background},
    literate=
     *{0}{{{\color{numb}0}}}{1}
      {1}{{{\color{numb}1}}}{1}
      {2}{{{\color{numb}2}}}{1}
      {3}{{{\color{numb}3}}}{1}
      {4}{{{\color{numb}4}}}{1}
      {5}{{{\color{numb}5}}}{1}
      {6}{{{\color{numb}6}}}{1}
      {7}{{{\color{numb}7}}}{1}
      {8}{{{\color{numb}8}}}{1}
      {9}{{{\color{numb}9}}}{1}
      {:}{{{\color{punct}{:}}}}{1}
      {,}{{{\color{punct}{,}}}}{1}
      {\{}{{{\color{delim}{\{}}}}{1}
      {\}}{{{\color{delim}{\}}}}}{1}
      {[}{{{\color{delim}{[}}}}{1}
      {]}{{{\color{delim}{]}}}}{1},
}

\begin{lstlisting}[language=json,firstnumber=1]
{"skill_name": "play music",
 "sample_utterances": [
  {"id": 0, "intent": "PlayMusicIntent", "text": "play {MusicName} please"},
  {"id": 1, "intent": "PlayMusicIntent", "text": "i want to listen to {MusicName}"}, 
  {"id": 2, "intent": "PlayMusicIntent", "text": "can you play {MusicName}"},
  {"id": 3, "intent": "PauseIntent", "text": "stop playing"},
  {"id": 4, "intent": "ResumeIntent", "text": "resume playing"}
  ],
  "slots": [{"name": MusicName, "values": ["shape_of_you", "frozen", "despacito"]}]
 }

\end{lstlisting}

\textbf{Equivalence sets of paraphrases.} To train our neural paraphrase model, we use an internal dataset of spoken utterances and the external public dataset PPDB \citep{ganitkevitch2013ppdb}. The internal data consists of a number of utterances in different domains and various skills that are manually annotated with intents and slots. Examples
for intents and slots are shown in Table \ref{tab:examples}. We define two utterances as
semantically equivalent if and only if they are annotated with the same
domain or skill, intent, and {\em set\/} of slots; we then say that these utterances have the same
\emph{signature}. This equivalence relation is considerably weaker than full meaning identity (since, for example, it does not take slot order into account), but practically useful nevertheless.

Further, when creating training data for paraphrasing, we
delexicalize utterances by replacing slot values with slot names; this allows
us to focus on syntactic variations rather than on slot values. Grouping  utterances by their \emph{signature}, as well as delexicalizing the slots (as illustrated in Table
\ref{tab:examples}), enables us to build large sets of paraphrases. In addition, since developers are required to add delexicalized grammar samples in ASK, our model can help to suggest possible utterances based on the examples developers provide during skill development stage.

The following are the main contributions of this paper: 
\begin{itemize}
	\itemsep-0.2em
	\item We use semantic equivalence classes based on the notion of \emph{signatures}. This relaxation of strict semantic equivalence advances the prior paraphrasing paradigm.
	\item We generate paraphrases of delexicalized utterances, utilizing slot information from back-propagating through the  values.
	\item We use pointers to copy slots which do not appear in the training data, thereby alleviating out-of-vocabulary problems during inference.
	\item We formally define various metrics to measure paraphrase quality, and use them to prove the effectiveness of the proposed sequence-of-sequence-of-sequence-to-sequence model and pointer network.
	\item We show that high-quality paraphrases that match live human utterances can improve downstream NLU tasks such as IC (intent classification) and NER (named entity recognition).
\end{itemize}

\begin{table}[t]
	\captionsetup{font=small}
	\centering
	\resizebox{14cm}{!}{
	  \begin{tabular}{|l|c|c|c|c} 
		\hline
		\textbf{Paraphrases?}	& \textbf{Utterance} & \textbf{Delexicalized Utterance} & \textbf{Signature}  \\
		\hline
		
		 Yes     & \pbox{20cm}{ Can you read me a book by \\ Shakespear about romance}
					& \pbox{20cm}{ Can you read me a book by \\ \{author\} about \{topic\} } 
					& \pbox{20cm}{ Books, ReadBook, \\ \{author\}, \{topic\} }\\
		\cline{2-3}
		 			& \pbox{20cm}{ Could you find a comic book \\ written by Mark Twain }
		 			& \pbox{20cm}{ Could you find a \{topic\} book \\ written by \{author\} }
		 			& \\
		\hline
		
		 No      & \pbox{20cm}{ What are the movies on show \\ near Seattle}
					& \pbox{20cm}{ What are the moves on show  \\ near \{location\} } 
					& \pbox {20cm}{ Cinema, FindMovie, \\ \{location\} }\\
		\cline{2-4}
		   	  		& \pbox{20cm}{  Find movies in Chicago area \\ on Saturday } 
		 			& \pbox{20cm}{  Find movies in \{location\} area \\ on \{date\} } 
		 			& \pbox {20cm}{ Cinema, FindMovie, \\ \{location\}, \{date\} } \\
		\hline
	  \end{tabular}
    } 
    \caption{Utterances with the same \emph{signature} are considered paraphrases of
      one another. Slot names are in curly brackets. The \emph{signature} of an utterance $u$ 
      consists of $u$'s domain, intent, and set of slots.}
	\vspace{-0.6em}
	\label{tab:1}
\label{tab:examples}\end{table}

\section{Related Work}

To the best of knowledge, our research is the first to generate delexicalized paraphrases by leveraging entity information directly within a neural network. \citet{malandrakis2019controlled} introduce a similar notion of paraphrasing and apply variational autoencoders to control the quality of paraphrases. \citet{sokolov2020neural} tackle a similar problem of paraphrasing utterances with entity types, but implement the slot copy mechanism via pre-processing and post-processing. In addition, \citet{liu2013paraphrase} apply paraphrases to improve natural understanding in an NLU system, both for augmenting rules and for enhancing features.

\section{Model}
\begin{figure*}[h!]
  \centering
  \captionsetup{font=small}
   \includegraphics[width=\linewidth]{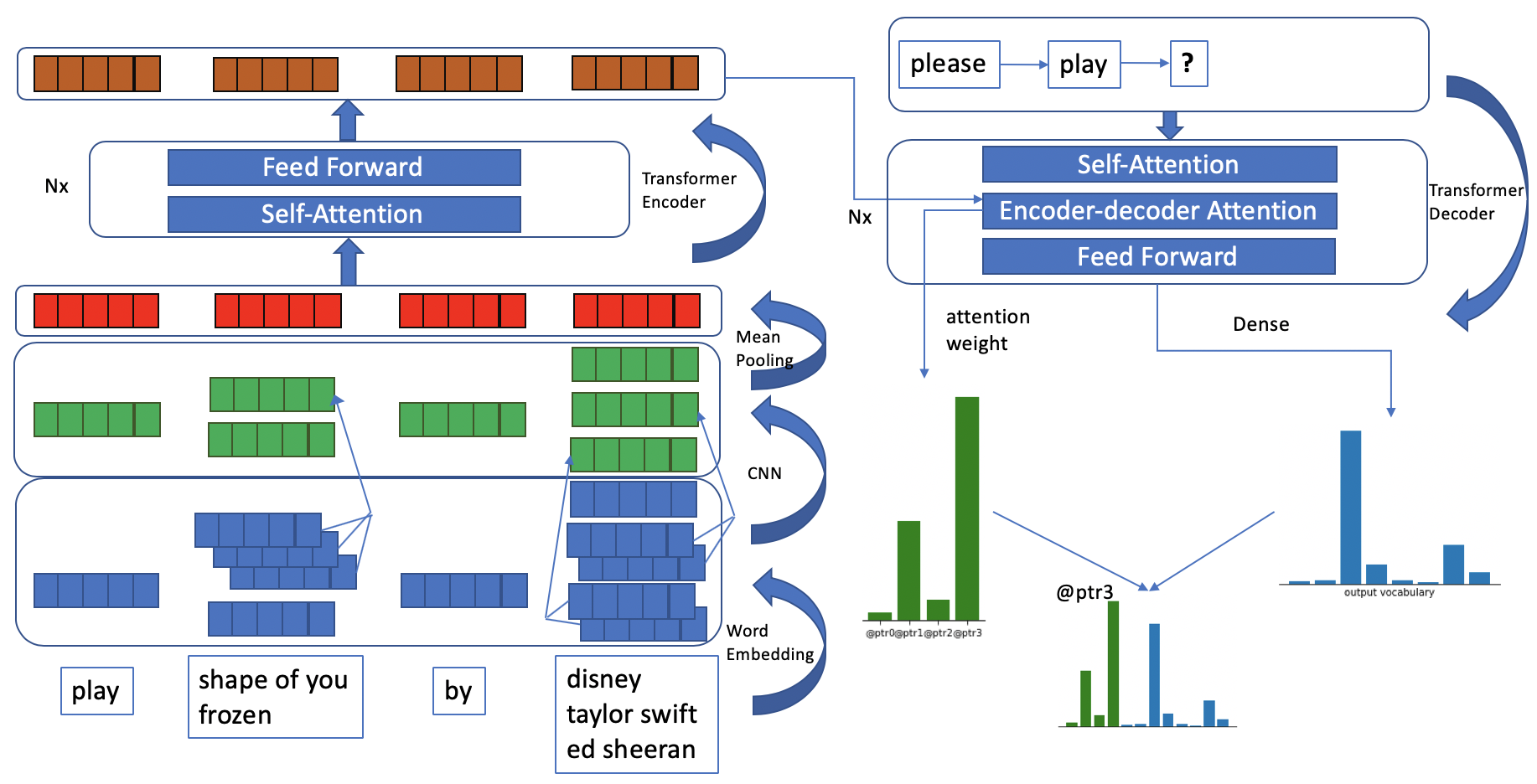}
   \caption{Sequence-of-Sequence-of-Sequence to Sequence Pointer Network to Generate Delexicalized Paraphrases}
  \label{fig:model}
\end{figure*}

We use the encoder-decoder sequence-to-sequence model \citep{sutskever2014sequence}.
The encoder embeds an input sentence via transformers \citep{vaswani2017attention}. The decoder is also a transformer model that generates output one token at a time, using information from the encoder and the previous time steps. An attention mechanism \citep{luong2015effective} helps the decoder to focus on appropriate regions of the input sentence while producing each output token. 
A good paraphrase should contain all the entity slots names from the source; some words remain the same in the paraphrase. To facilitate such copies, we use pointers that directly copy input tokens to the output \citep{rongali2020don}. As a result, in cases where an input token does not exist in the vocabulary, the model will learn to make the copy based on its embedding and context. Figure \ref{fig:model} depicts our proposed architecture. 

\subsection{Input Embedding}
One of the biggest challenges in our paraphrasing problem is how to deal with slots in sentences. Slots can come in a variety of flavors: 
\begin{itemize}
\item Well-defined and popular slots (like music names, city names, or numbers).
\item Partially defined slots, like a \textit{Horoscope} slot that has a few samples (such as \textit{Leo}, \textit{Aquarius} and \textit{Sagittarius}).
\item Free-form slots that may include any random values.
\end{itemize}
Across different domains and skills, we might see slots from all three categories. 

Similar to traditional sequence models for generation, we start with directly using delexicalized utterances in input and output, like ``find movies in \{location\} on \{date\}'' in the example above. Notice here that \{location\} and \{date\} define entity slots that may not have general semantics. We have observed that in the case of skills, each skill may have its own specific slots, and thus we may see millions of different tokens for slot values. There is little information to be gained by learning each slot value, and during inference we might see out-of-vocabulary tokens often. This model uses a direct sequence embedding layer, and we refer to it as S1. Also, at a later stage, in order to generate unseen slots during inference, we will need to anonymize the slot, and that will be referred to as AS. In this case, all slots with be replaced by tokens SLOT1, SLOT2, ... etc. in the order of their occurrence in the sequence.

We propose an S2 embedding layer (sequence-of-sequence) and an S3 embedding layer (sequence-of-sequence-of-sequence)  for better handling of slots. 

In S2, each token in the input sequence can also be treated as a sequence of words. For example, a delexicalized utterance such as ``find movies in \{location\} on \{date\}'' can be rewritten as ``find movies in \textit{boston,new york} on \textit{tomorrow,march twenty first}''. The embedding of each token will simply be the average word embedding from the sub-sequence.

However, S2 may not solve our problem in all cases. Any slot value itself can also be another sequence of multiple words, as in ``find movies in \textit{boston,new york} on \textit{tomorrow,march twenty first}''. Phrase embeddings can be used  here instead of word embeddings, treating \textit{new york} or \textit{march twenty first} as single token. Alternatively, we add an extra convolutional layer on the sub-sub-sequence, and that will be S3. The 1D convolutional layer has a kernel size of 3, 512 channels and is followed by a dense layer to generate the phrase embedding.

In all cases, gradient descent will back-propagate all the way back to the average pooling layer and the convolutional layer. As a result, our model will learn to capture information from different slot values in any slot, and also to understand complex slot values.

\subsection{Transformer Encoder and Decoder}

We use the traditional transformer encoder and decoder for the seq2seq model. The embedding layer that maps input and output tokens to a vector is as defined in the previous section, while we have three different options to extract information from delexicalized sequences: AS, S2 and S3. Positional embedding will also be applied in the same way as in the original transformer model. 

Afterwards, the encoder is composed of a stack of identical layers, where each layer has two
sub-layers: a multi-head self-attention layer, and  a simple, fully connected layer. A residual connection is employed around each of the two sub-layers, followed by layer normalization. 

The decoder will be mostly similar to the encoder, including a multi-head self-attention layer and a fully connected layer, and also a third layer for multi-head attention over the encoder output.

\subsection{Pointer Network}

During the decoding stage, at each time step $t$ the transformer decoder generates a hidden-state vector $d_t$. By multiplying that vector with the output word embedding, we get a score for each word in the vocabulary $[s_1,...,s_{|V|}]$. A following softmax layer generates the probability for each word in vocabulary to be generated. Recall that in our case we are trying to paraphrase delexicalized utterances, where we sometimes need to generate slot names that might be out-of-vocabulary tokens. Previously, we applied a convolutional layer and mean pooling on the word embeddings, and managed to handle the problem in the encoder stage. However, similar technique cannot be directly applied in the decoder. 

Alternatively, we use pointers to implement a copy mechanism that can directly copy tokens in the input to the output. From the attention over the encoder we can get a score for each token in the input, indicating the strength of the relationship between that input token and the next time step token in the decoder, $[a_1,...,a_n]$. We concatenate the attention scores with the original unnormalized word scores, leading to a vector of $n+|V|$ dimensions $[a_1,...,a_n,s_1,...,s_{|V|}]$. The first $n$ items represent @ptr$i$ ($i=1,2,..,n$) tokens as in Table~\ref{tab:reformat}, indicating scores for each input token to be copied, and the rest are scores for the output vocabulary. We then apply a softmax layer, and the model will learn that either an input token is copied or an in-vocabulary word is generated. The application of a pointer network along with AS, S2 and S3 input embeddings is referred to as ASP, S2P and S3P, respectively. And the AS embedding can be applicable even without points, because slots are anonymized.

\subsection{Reformat data} \label{reformat_data}

For implementing the model described above, we modify both source and target data to include necessary information. Within the output of models with pointers, we use token @ptr$n$ to indicate that this token is directly copied from the $n$th token in the input. You can find an example in Table~\ref{tab:reformat}. 

\begin{table}[]
\captionsetup{font=small}
\centering
\resizebox{14cm}{!}{
\begin{tabular}{|l|l|l|l|}
\hline
Format  & Input                                                                                                & Output                                              & Slots                                                                                                                           \\ \hline
O       & play \{MusicName\} by \{ArtistName\} please                                                          & i want to listen to \{ArtistName\} 's \{MusicName\} & \multirow{4}{*}{\begin{tabular}[c]{@{}l@{}}MusicName:\\ frozen, shape of you\\ ArtistName:\\ taylor swift, disney\end{tabular}} \\ \cline{1-3}
AS      & play SLOT1 by SLOT2 please                                                                           & i want to listen to SLOT2 's SLOT1                  &                                                                                                                                 \\ \cline{1-3}
ASP     & play SLOT1 by SLOT2 please                                                                           & i want to listen to @ptr3 's @ptr1                  &                                                                                                                                 \\ \cline{1-3}
S2P/S3P & \begin{tabular}[c]{@{}l@{}}play frozen,shape\_of\_you \\ by taylor\_swift,disney please\end{tabular} & i want to listen to @ptr3 's @ptr1                  &                                                                                                                                 \\ \hline
\end{tabular}
}
 \caption{Different formats of training data}
 \small{O: Original. AS: Anonymized Slots. ASP: Anonymized Slots with Pointers. }
 
 \small{S2P: Sequence-of-Sequence with Pointers. S3P: Sequence-of-Sequence-of-Sequence with Pointers. }
	\label{tab:2}
	\label{tab:reformat}
\end{table}

\section{Experimental Setup} \label{sec:trainandinfer}
In this section we introduce the training and evaluation datasets we used, and the deep sequence-to-sequence model training environment.

\textbf{Dataset} We train paraphrase models using data from 58,000 skills, live non-skill utterances from broader domains, and the public dataset PPDB. We then apply these models to the 88 most popular skill in order to obtain paraphrases and calculate evaluation metrics. 

We generate 6.4 million paraphrase pairs from skills, which form the bulk of the training dataset. We also create another training set by appending an extra 500,000 non-skill paraphrase pairs, and two million pairs from PPDB. However, model performance here is not as remarkable. The public dataset PPDB we used only includes lexicalized paraphrase pairs, and those are generally sentences from the web and from various documents, which are a little different from our use case. In our task, including the public dataset does not seem to provide much extra gain. Thus, the discussion below focuses the analysis on results from the skill-only dataset.

Utterances are delexicalized, and each \emph{signature} (as defined in Section~\ref{sec:intro}) corresponds to a set of delexicalized utterances. 
We create two source-target pairs for each utterance, by randomly sampling its target from the same set. When training the model with pointers, slots in the target are replaced by respective pointers. We also cleaned up noisy data, so that utterances have reasonable length and contain enough contextual words around entities. 

As described in Section~3.4, the training dataset is reformatted into four different types, with various paraphrasing models trained on each of them.

\textbf{Training and Inference} 
We implemented the special input embedding layer and transformers with pointers in MXnet 1.5.0 (Gluon API). All models are trained with the same hyperparameters for fair comparison.
Both the transformer encoder and the decoder include 8 heads, 6 layers, a hidden size of 512, and a 0.1 dropout \citep{srivastava2014dropout} ratio. 
The Adam optimizer \citep{kingma2014adam} and Noam learning rate scheduler are used, with an initial learning rate of 0.35 and 4000 warm-up steps. The model is trained for 40 epochs with a batch size of 1400 and with 8 batches per update.
Inference is performed with a beam-search decoder. The beam size is 5 and 3-best paraphrases are kept for each input. 


\section{Evaluation}\label{sec:evaluation}

There are numerous evaluation metrics for sequence generation problems, such as BLEU \citep{papineni2002bleu} and ROUGE \citep{lin2004rouge}. However, in our case we do not have ground truth for paraphrases and thus it would be hard to directly apply these metrics. We now describe how to evaluate paraphrase generation for the use case of data augmentation, and propose several intrinsic metrics that emphasize different characteristics. We hypothesize that paraphrases which benefit downstream models should have the following properties: divesity, novelty, and good coverage of test data. We describe each of these in detail below. 

\subsection{Intrinsic Metrics}
We use $\mathcal{D}$ to denote the set of delexicalized utterances available at training time, and $G(\mathcal{D})$ to denote the set of generated paraphrases.

\textbf{Slot Copy Error Rate} calculates the ratio of slot copy misalignment in the generation. In some cases, not all slots in the input are copied into the output. To calculate this metric, all sample utterances in our 88-skill dataset are run through the paraphrasing model, generating paraphrased utterances for each; we then measure the fraction of generated utterances that don't match the source utterance slots. This metric indicates how well the model is able to identify and copy all slots in the source sequence.

\textbf{Novelty} is the proportion of generated utterances which are not in original paraphrase sets. This metric should give an indication of how much paraphrasing can be expected to help in augmenting grammar samples and training data: 
$$ \frac {|G(\mathcal{D}) \setminus \mathcal{D}|} {|G(\mathcal{D})|}$$

\textbf{Diversity} is the number of unique generated utterances: 
$$|G(\mathcal{D})| $$

\textbf{Trigram Novelty and Trigram Diversity} We notice that many generated paraphrases are  minor modifications of existing utterances, e.g., obtained by inserting or removing stopwords
like ``the'' or ``please.'' To gauge the ability of the paraphrasing model to generate sequences with larger structural differences (like creating a passive voice from an active voice), in addition to metrics at utterance level we also evaluate novelty and diversity at trigram level. This metric is similar to an inverse ROUGE-3 metric between input sequences and paraphrase outputs.

\subsection{Extrinsic Metrics}
In this paper we also consider downstream NLU applications, including IC and NER. We use the Alexa Skills Kit base pipeline \citep{kumar2017just}, which builds NLU models from delexicalized utterances and slots. The model includes Finite-State Transducers (FSTs) for capturing exact matches and a DNN statistical model on joint IC/NER tasks. The network consists of shared bi-LSTM layers from pre-training, skill-specific bi-LSTM layers, and on top of those two individual branches it features a dense layer and a softmax for IC,  along with a dense layer and a CRF layer for NER.


For each skill, the FST is constructed from delexicalized samples and slot value samples. For the statistical model,  training data is sampled from delexicalized utterances. During lexicalization, each slot is replaced with a word or phrase uniformly sampled from its entity list. We apply paraphrasing models to delexicalized samples, and augment both the FST and the DNN model training data. The added samples will first go through an intent classification filter by filling the slots and predicting the intent using the original model, and then only samples which retain the intent are added for data augmentation.

Finally, each model is applied on test data  and we calculate the following metrics:

\textbf{Intent Filter Rate} evaluates the proportion of paraphrases which belong to the same intent.

\textbf{FST New Rules} is the total number of delexicalized samples added in all skills. The samples serve both as additional FST rules and as extra training data for statistical models.

\textbf{FST New Matches} is the percentage of live utterances in the test data that are matched by FSTs. This metric measures whether the generated paraphrases capture what users say exactly.

\textbf{Intent Error Rate} measures accuracy in the intent classification model. It is the proportion of utterances where the model makes an intent error.

\textbf{Slot Error Rate} is a metric for evaluating NER. It is defined as $$\textrm{SER} = \frac {S+I+D} {\textrm{Total number of slots}}$$ where $S$, $I$ and $D$ are the numbers of substituted, inserted, and deleted slots, respectively.


\textbf{Semantic Error Rate} \citep{makhoul1999performance} is a joint metric for both IC and NER. It is defined as $$\textrm{SEMER} = \frac {S+I+D+IE} {\textrm{Total number of slots} + 1}$$ where $S$, $I$ and $D$ are again the numbers of substituted, inserted, and deleted slots, respectively. $IE$ is 1 if there is an intent error and 0 if the model predicts the correct intent.

\section{Results}
The intrinsic and extrinsic results are presented in Table \ref{tab:intrinsic} and Table \ref{tab:extrinsic}. In Table \ref{tab:example} you can find examples of paraphrases from three different skills.

\begin{table}[]
\captionsetup{font=small}
\centering
\resizebox{12.5cm}{!}
{
\begin{tabular}{c|c|c|c|c|c}
Model & Slot Copy Rate & Novelty & Diversity & Trigram Novelty & Trigram Diversity \\ \hline
AS    & 99.98\%              & 72.21\% & 33408     & 59.27\%         & 39429             \\
ASP   & 92.22\%              & 53.70\% & 36824     & 48.35\%         & 41614             \\
S2P   & 84.90\%              & 53.42\% & 35315     & 45.95\%         & 39470             \\
S3P   & 87.47\%              & 54.16\% & 35706     & 47.88\%         & 36779            
\end{tabular}
}
    \caption{Intrinsic metrics from different models}
	\label{tab:intrinsic}
\end{table}

\begin{table}[]
\captionsetup{font=small}
\centering
\resizebox{14cm}{!}
{
\begin{tabular}{c|c|cc|ccc}
    & \multicolumn{1}{l|}{}                   & \multicolumn{2}{c|}{FST} & \multicolumn{3}{c}{NLU}                    \\ \hline
    & \multicolumn{1}{l|}{Intent Filter Rate} & New Rules  & New Matches & Intent Error & Slot Error & Semantic Error \\ \hline
AS  & 54.31\%                                 & 13103      & 1528        & -2.75\%      & -7.97\%    & -3.65\%        \\
ASP & 61.87\%                                 & 12235      & 1669        & -2.58\%      & -0.74\%    & -2.28\%        \\
S2P & 58.21\%                                 & 10982      & 1376        & -0.20\%      & -0.93\%    & -0.83\%        \\
S3P & 60.04\%                                 & 11610      & 1438        & -1.60\%      & -3.02\%    & -0.78\%       
\end{tabular}
}
    \caption{Extrinsic metrics from different models}
    \small{Note: All NLU metrics are relative numbers because we cannot disclose absolute numbers}
	\label{tab:extrinsic}
\end{table}

\textbf{Intrinsic Metrics}
The usual sequence-to-sequence model achieves the highest novelty, while all models with pointers have similar numbers. We also investigate how likely it is that the paraphrase has the same set of slots with the source, which aligns to our definition of \emph{signature}. (Note that utterances with no slots are not included when calculating this metric.) Overall, using pointers in the decoder stage does not benefit slot copying: When all slots are anonymized, a usual sequence decoder gives near perfect slot copy rate, because anonymized slot names like SLOT1 and SLOT2 provide direct strong signals indicating that this token is extremely likely to be copied, both in the encoder and in the decoder. The pointer decoder gives 92.22\%, and we find in most misalignment cases the output sequence misses a pointer to a slot in the input. Pointers compete with the vocabulary from the final softmax in the decoder and they rely on context (encoder state) to identify the logit, which may bring much noise, and hence it's reasonable to see the relatively lower copy rate in the AS case of simply generating very frequent SLOT1, SLOT2 output tokens. 

\begin{figure}[h!]
  \centering
  \captionsetup{font=small}
   \includegraphics[scale=0.6]{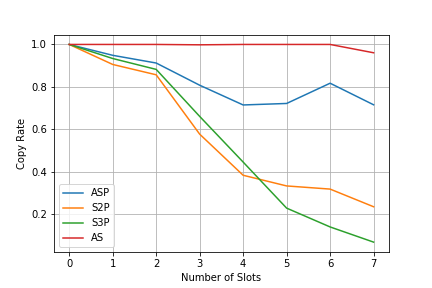}
   \caption{Variation of Slot Copy Rate by Number of Slots}
  \label{fig:slot}
\end{figure}

For S2P and S3P, the tokens to be copied are unknown and have embeddings originating from extra layers. We still see over 80\% copy rate, and the chance of perfectly copying all the slots reduces as the number of slots increases. As is shown in Figure \ref{fig:slot}, the exact copy rate is greater than 90\% for all pointer models if there is only one slot in the input, while the number for vanilla seq2seq model is 100.0\%. The proportion of exact slot copy falls drastically as the number of slots increases, especially for the S2P and S3P models. For the ASP model, the explicit token SLOTX in the input provides an indication for it to be copied, but in S2P and S3P the token could be from the pooling of embeddings from various words and makes it hard for the pointer to locate all slots, thus we see a sharper decrease of copy rate. In future work, in order to improve the copy rate of pointer models, we can try to add extra signals to the input, indicating whether each token is a slot, as well as an extra connection between the input sequence and the decoder.

For the novelty and diversity metrics, both at utterance level and trigram level, there is not much difference among models with pointers. Vanilla seq2seq model with anonymized slots generates the most unique utterances. However, since no information on slot values is provided, some of the generations may not be proper paraphrases. As in our definition of paraphrases, the generated utterance must belong to the same intent. In the next section on extrinsic metrics, we can see that vanilla seq2seq is least likely to retain the intent.

From the internal metrics, we might not see benefits from pointer decoders, as they limit the span of generation and do not copy as much slots. However, novel generated utterances might be just random and do not possess similar semantics, even if same slots are included. In addition, for models with anonymized slots, since the slot tokens do not convey any information about semantics, we expect to see more natural and proper generation from the S2P and S3P models.

\textbf{Extrinsic Metrics}
Overall, our paraphrasing models generate utterances that help both FST matching and NLU models. Within 129,599 test utterances, we see 1,669 new FST matches in the best model. Paraphrasing as data augmentation also benefits both IC and NER, leading to a reduction in slot error rate, intent error rate and semantic error rate.

Models with pointers all show higher intent filter rate, suggesting that the direct connection to encoder output helps the decoder to locate appropriate slots in the input, and consequently the context words learn from the generated pointer and eventually generate a sequence with more similar semantics.

The number of new FST matches is an essential metric for evaluating the quality of paraphrases, as they demonstrate whether the model can learn what humans are likely to say. From various data sources, especially enormous numbers of diversified skills, our model learns to gather information from similar skills and also adapts to what people usually say when using a virtual assistant. All models generate considerable numbers of new FST matches: Out of 40,109 utterances that were not matched by the original FSTs,  the number of new matches from AS, ASP, S2P and S3P models are 1,528, 1,669, 1,376 and 1,438, respectively. Anonymized slots with pointers achieve the highest number of exact matches, which further highlights the effectiveness of pointers. 

We also see improvements of downstream NLU tasks by applying paraphrases to data augmentation. After lexicalizing generated paraphrases, those are filtered by intent and then added to the training data for IC and NER, and evaluated on a multi-task DNN model with bi-LSTM encoders and decoders. We calculate evaluation metrics mentioned in section 4.2 for all 88 skills, and report the average. 

Adding extra paraphrases improves both IC and NER for all our proposed paraphrasing models. AS, the most naive model which anonymizes slots and does not use pointers, achieves the highest performance. We see a 2.75\% relative reduction in intent error rate, 7.97\% reduction in slot error rate and 3.65\% reduction in semantic error rate.

\textbf{Skill Analysis} 
When using paraphrases for data augmentation using the AS model, among all 88 skills, 44 see improvement in SEMER, with 15 improving by more than 2\%; 34 see degradation, with 7 of them degrading by more than 2\%. We want to understand what kind of skills benefit most from paraphrases. For investigating such skill characteristics, we calculate Spearman's rank-order correlation coefficient between SEMER relative improvements and skill features including number of intents, number of slots, number of unique delexicalized utterances, number of unique lexicalized utterances, number of unique delexicalized utterances per intent, and number of lexicalized utterances per intent. The highest correlation is between SEMER improvement and the number of unique delexicalized utterances, with Spearman correlation coefficient -0.240 and \textit{p}-value of 0.024, indicating that our paraphrasing model will benefit more for skills with scarce delexicalized samples.

\begin{table}[]
\captionsetup{font=small}
\centering
\resizebox{16cm}{!}
{
\begin{tabular}{|llll|}
\hline
\multicolumn{1}{|l|}{Original Utterances}                                                                                                                                                                                             & \multicolumn{1}{l|}{Paraphrases from both AS and S3P}                                                                                             & \multicolumn{1}{l|}{Paraphrases from AS only}                                                                                                                                                                                                      & Paraphrases from S3P only                                                                                                                                                                                                                                                   \\ \hline
\multicolumn{1}{|l|}{\begin{tabular}[c]{@{}l@{}}\{comedian\} show\\ play \{comedian\} stand up\\ \{comedian\} performance\\ \{comedian\}\\ to play \{comedian\}\\ search for \{comedian\}\\ about \{comedian\} stand up\end{tabular}} & \multicolumn{1}{l|}{\begin{tabular}[c]{@{}l@{}}find \{comedian\}\\ tell me about \{comedian\}\\ the \{comedian\} show\end{tabular}}              & \multicolumn{1}{l|}{\begin{tabular}[c]{@{}l@{}}what is \{comedian\}\\ a \{comedian\} standup comedy\\ \{comedian\} up\\ play \{comedian\} game\\ i think it is \{comedian\}\\ play a standup of \{comedian\}\\ play the \{comedian\}\end{tabular}} & \begin{tabular}[c]{@{}l@{}}i want to hear \{comedian\}\\ listen to \{comedian\}\\ play \{comedian\} standup comedy\\ to play \{comedian\} podcast\\ what is the performance of \{comedian\}\\ what is the status of \{comedian\}\\ what is \{comedian\} doing\end{tabular} \\ \hline
                                                                                                                                                                                                                                      &                                                                                                                                                  &                                                                                                                                                                                                                                                    &                                                                                                                                                                                                                                                                            \\ \hline
\multicolumn{1}{|l|}{\begin{tabular}[c]{@{}l@{}}play noise \{item\}\\ play \{item\}\\ play sound \{item\}\\ play \{item\} sounds\\ play song \{item\}\\ play the song \{item\}\end{tabular}}                                          & \multicolumn{1}{l|}{\begin{tabular}[c]{@{}l@{}}play \{item\} song\\ play \{item\} sound\end{tabular}}                                            & \multicolumn{1}{l|}{\begin{tabular}[c]{@{}l@{}}make the \{item\} screensaver\\ make the \{item\} sound\\ take the sound number \{item\}\\ \{item\} sounds\end{tabular}}                                                                            & \begin{tabular}[c]{@{}l@{}}play ambient sound \{item\}\\ play \{item\} please\\ sing song \{item\}\end{tabular}                                                                                                                                                            \\ \hline
                                                                                                                                                                                                                                      &                                                                                                                                                  &                                                                                                                                                                                                                                                    &                                                                                                                                                                                                                                                                            \\ \hline
\multicolumn{1}{|l|}{\begin{tabular}[c]{@{}l@{}}i'd rather \{answer\}\\ i would rather \{answer\}\\ probably \{answer\}\\ i choose \{answer\}\\ maybe \{answer\}\\ i would \{answer\}\end{tabular}}                                   & \multicolumn{1}{l|}{\begin{tabular}[c]{@{}l@{}}i think it's \{answer\}\\ i think it is \{answer\}\\ i would like to try \{answer\}\end{tabular}} & \multicolumn{1}{l|}{\begin{tabular}[c]{@{}l@{}}i \{answer\}\\ is it \{answer\}\end{tabular}}                                                                                                                                                       & \begin{tabular}[c]{@{}l@{}}a \{answer\}\\ i want \{answer\}\end{tabular}                                                                                                                                                                                                   \\ \hline
\multicolumn{1}{|l|}{\{answer\}}                                                                                                                                                                                                      & \multicolumn{1}{l|}{N/A}                                                                                                                         & \multicolumn{1}{l|}{i think it is \{answer\}}                                                                                                                                                                                                      & the \{answer\}                                                                                                                                                                                                                                                             \\ \hline
\end{tabular}
}
    \caption{Paraphrase Examples}
	\label{tab:example}
\end{table}

\textbf{Paraphrase Examples} Among the three paraphrasing examples shown in Table \ref{tab:example}, each  behaves differently on the SEMER evaluation metric. The first row is from the skill where data augmentation from all models outperform the baseline, and the S3P model greatly outperforms the AS model. The second row is from the skill where AS greatly outperforms S3P. The third and fourth rows are utterances of two different intents from a skill where all data augmentation techniques degrade  NLU performance. 

The first example is from a skill for comedian shows. Alexa users can ask to play a comedian's show or to search for comedians. As is shown in the examples, the S3P model learns from the CNN and embeddings of artists' names, and understands that \{comedian\} is a person, thus generates utterances like ``i want to hear \{comedian\}'' and ``what is \{comedian\} doing.'' In contrast, the model with anonymized slots treats \{comedian\} as a general slot without any extra information, and as a result generates paraphrases that are not appropriate for this skill, like ``what is \{comedian\}'' and ``play \{comedian\} game.''

The second example is from a skill for playing different kinds of sounds. From the examples, it is apparent that S3P is generating better paraphrases. However, downstream NLU tasks perform better with paraphrases from AS. The shown examples are sample utterances for PlaySoundIntent, however, there is another PlayAmbientSoundIntent in the skill. Notice that S3P generates a paraphrase ``play ambient sound \{item\}'' and probably due to a defect of intent filtering, the utterance is not filtered out. After the paraphrase is added to the training data, the statistical model may get confused on similar utterances for playing ambient sounds.

The third example shows utterances for two different intents, AnswerIntent and AnswerOnlyIntent. The skill intends to create a class for \{answer\} without any carrier phrases. However, the paraphrasing models have no knowledge of this objective and generate utterances by adding context words. The original intent classfication cannot filter out all of these cases. And afterwards, adding these samples to the FST and training data will further confuse the NLU model.

Overall, downstream NLU tasks may not be best indicators for paraphrase quality. S3P models show the effectiveness of incorporating entity values knowledge in paraphrase generation, which may or may not lead to an accuracy gain on downstream NLU tasks. Some heavy manual evaluations might provide a more accurate overview for comparison among different paraphrasing models.

\section{Conclusion}

 We trained and evaluated multiple types of models for paraphrasing delexicalized utterances, motivated to assist skill developers and ultimately to improve the user experience of virtual assistant customers. We experimented with anonymizing entity slots in utterances, applying CNNs and pooling on slot entities, and using pointers to locate slots in the output. The generated paraphrases bring about 1,669 exact matches with human utterances in the best model, and also improve NLU tasks, especially for those skills with insufficient training samples. In addition, we showed the benefit of including slot value information in paraphrasing for some skills. 


\bibliographystyle{acl_natbib}
\bibliography{coling2020}

\end{document}